\documentclass[10pt,twocolumn,letterpaper]{article}

\usepackage{iccv}
\usepackage{times}
\usepackage{epsfig}
\usepackage{graphicx}
\usepackage{amsmath}
\usepackage{amssymb}

\usepackage{algorithm}
\usepackage{algpseudocode}
\usepackage{multirow}
\usepackage{tabularx,verbatim}
\usepackage{comment}
\usepackage{bbm}
\usepackage{enumitem}
\usepackage{colortbl}
\usepackage{color, xcolor} % colors
\usepackage{bbding}
\usepackage{booktabs}
\usepackage{dsfont}
\usepackage[accsupp]{axessibility}  % Improves PDF readability for those with disabilities.

\newcommand{\ytComment}[1]{\textcolor{black}{#1}}

\usepackage[breaklinks=true,bookmarks=false]{hyperref}

\usepackage{xcolor}
\hypersetup{
    colorlinks=true,    % false: boxed links; true: colored links
    citecolor=green, %OrangeRed, % green,    % color of links to bibliography
    filecolor=magenta,  % color of file links
    urlcolor=blue       % color of external links
}

\iccvfinalcopy % *** Uncomment this line for the final submission

\def\iccvPaperID{2338} % *** Enter the ICCV Paper ID here

\ificcvfinal\fi

\begin{document}

%%%%%%%%% TITLE
\title{Generalized Few-Shot Point Cloud Segmentation Via Geometric Words}

% \author{Yating Xu\\
% National University of Singapore\\
% {\tt\small xu.yating@u.nus.edu}
% \and
% Conghui Hu\\
% National University of Singapore\\
% {\tt\small conghui@nus.edu.sg}
% \and
% Gim Hee Lee\\
% National University of Singapore\\
% {\tt\small gimhee.lee@nus.edu.sg}
% \and
% Na Zhao\\
% Singapore University of Technology and Design\\
% {\tt\small na\_zhao@sutd.edu.sg}
% }
\newcommand*\samethanks[1][\value{footnote}]{\footnotemark[#1]}

\author{
Yating Xu$^1$ \qquad Conghui Hu$^1$ \qquad Na Zhao$^2$\thanks{Na Zhao was concurrently a visiting professor at the National University of Singapore when this work was done.} \qquad Gim Hee Lee$^1$
\\
$^1$Department of Computer Science, National University of Singapore\\
$^2$Singapore University of Technology and Design\\
{\tt\small xu.yating@u.nus.edu \qquad conghui@nus.edu.sg \qquad na\_zhao@sutd.edu.sg \qquad gimhee.lee@nus.edu.sg}
}

\maketitle
\ificcvfinal\thispagestyle{empty}\fi

%%%%%%%%% ABSTRACT
\begin{abstract}
Existing fully-supervised point cloud segmentation methods suffer in the dynamic testing environment with emerging new classes.
Few-shot point cloud segmentation algorithms address this problem by learning to adapt to new classes at the sacrifice of segmentation accuracy for the base classes, which severely impedes its practicality. This largely motivates us to present the first attempt at a more practical paradigm of generalized few-shot point cloud segmentation, which requires the model to generalize to new categories with only a few support point clouds and simultaneously retain the capability to segment base classes.
We propose the geometric words to represent geometric components shared between the base and novel classes, and incorporate them into a novel geometric-aware semantic representation to facilitate better generalization to the new classes without forgetting the old ones.
Moreover, we introduce geometric prototypes to guide the segmentation with geometric prior knowledge.
Extensive experiments on S3DIS and ScanNet consistently illustrate the superior performance of our method over baseline methods. Our code is available at: \url{https://github.com/Pixie8888/GFS-3DSeg_GWs}.
% Our code will be open-sourced upon paper acceptance.
%We conduct various settings on S3DIS and ScanNet to verify the effectiveness of our method.
\end{abstract}

\section{Introduction}
Point cloud segmentation aims at predicting the category of each point in the 3D scenes represented by the point cloud, and has wide applications in autonomous driving, robotics, \etc. Although fully-supervised point cloud segmentation methods (Full-3DSeg) \cite{qi2017pointnet, qi2017pointnet++, wang2019dynamic, li2018pointcnn, lai2022stratified, wang2022semaffinet} have achieved impressive performance, they heavily require large-scale annotated training data and rely on the closed set assumption that the class distribution of the testing point cloud remains the same as the training dataset. However, the closed set assumption is unrealistic in the open world where new classes arise continuously. 
The Full-3DSeg in this challenging open-world setting thus requires large amounts of annotated data for new classes, which are time-consuming and expensive to collect. 
%Moreover, even with the availability of the annotated data for the new classes, Full-3DSeg would suffer from catastrophic forgetting of the base classes when directly fine-tuned without accessing training data of the base classes.

%Full-3DSeg can only be able to segment new classes when sufficient training samples are provided, but in the meantime, they are prone to catastrophic forgetting of the base classes.

% they cannnot be directly applied to unseen classes unless sufficent training samples of the target class are provided. Even though the large number of training data is provided, fine-tuning on novel classes requires large computation and has the risk of catastrophic forgetting.

% To ameliorate the lack of data for novel class adaptation, the few-shot point cloud segmentation (FS-3DSeg) \cite{zhao2021few, mao2022bidirectional, lai2022tackling} algorithms are designed.
The few-shot point cloud segmentation (FS-3DSeg) \cite{zhao2021few, mao2022bidirectional, lai2022tackling} algorithms are designed to ameliorate the lack of data for novel class adaptation.
Generally, FS-3DSeg first trains a model with abundant training samples of the base classes, and then targets at segmenting the new classes by learning from only a small number of samples from the corresponding new classes. 
By adopting episodic training \cite{vinyals2016matching} to mimic the testing environment and specific designs for feature extraction \cite{zhao2021few}, FS-3DSeg achieves promising novel class segmentation results for the query point clouds. 
However, in the task of point cloud segmentation, base and novel classes often appear together in one scene (see Figure \ref{s3dis-seg} as examples). An ideal segmentor in practice is expected to give each point in the scene a semantic label. %As such
As a result, the FS-3DSeg setting that only segments points of novel classes while ignoring the %points of 
base classes suffers limited practicality. %greatly limits its practical value.

\begin{figure}[t]
\centering
\includegraphics[scale=0.6]{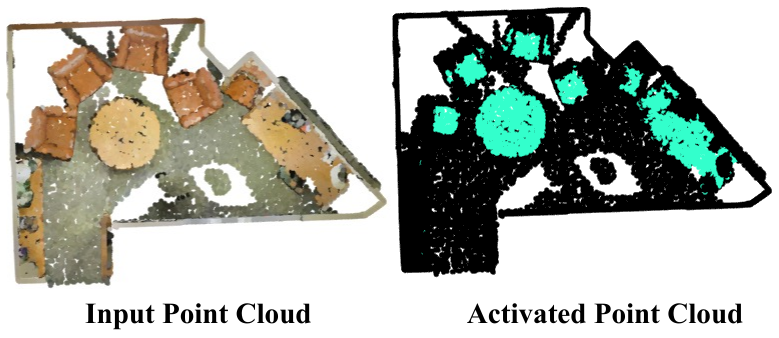}
\caption{\small{\textbf{Visualization of the geometric words (GWs) on S3DIS}. The left figure shows the original point cloud and the right figure shows the activated point cloud to a geometric word. The activated points are colored green. It shows the horizontal planes of the sofa, table and chair are all activated to the same GW due to similar geometric structure.}}
\label{visualize-gw}
\vspace{-3mm}
\end{figure}

\ytComment{
%To this end
In view of the impracticality of FS-3DSeg, we introduce the generalized few-shot point cloud segmentation (GFS-3DSeg) task. As shown in Tab.~\ref{setting comparison}, given the model originally trained on base classes, the objective of GFS-3DSeg is to segment both base and novel classes using merely a limited number of labeled samples for the new classes during %the 
testing. %stage.
%the objective of GFS-3DSeg is to segment both base and novel classes during the testing stage. And it is intended to achieve this by leveraging just a few labeled samples for the new classes.
% Notably, we restrict the model's access to the base dataset $D_\text{train}^b$ during the testing stage, accounting for concerns like data privacy and storage memory limitations. This exacerbates challenges faced by GFS-3DSeg. 
Furthermore, there is no access to the base training data %$D_\text{train}^b$ 
during testing in GFS-3DSeg considering the issues of data privacy and storage memory limitations. %, which are good for practical applications with data privacy and storage memory limitations. %This exacerbates challenges faced by GFS-3DSeg. 
%In our proposed setting, we take into account common challenges like data privacy and limited memory storage. This leads us to  restrict the model's access to the base dataset during the testing stage, which further exacerbate the difficulties encountered by GFS-3DSeg. 
%Consequently
In this demanding but practical context, we expect a good generalized few-shot point cloud segmentor to effectively learn to segment novel classes with few samples and also maintain the knowledge of the base classes. 
A potential solution to GFS-3DSeg %might adopt
can be using prototype learning \cite{tian2022generalized, qi2018low} to generate class prototypes of the base and novel classes as the classifier weights, which can quickly adapt to new classes and %also 
avoid forgetting past knowledge caused by fine-tuning.
% A straightforward solution to GFS-3DSeg might adopt the non-parametric classifier as in prototypical learning \cite{snell2017prototypical}, where the prototypes of base classes are preserved to avoid catastrophic forgetting of the past knowledge and the prototypes of novel classes are learned from the support sample. 
However, effective learning of the representation and classifier for the novel classes is not trivial and remains as a major challenge.
}

In this paper, we propose \textbf{geometric words}\footnote{Analogous to visual words in bag-of-words image retrieval systems \cite{nister2006scalable}.} (GWs) as the transferable knowledge obtained from the base classes to enhance the learning of the new classes without forgetting the old ones. Although different (old and new) classes contain distinct semantic representations, they usually share similar local geometric structures as shown in Fig.~\ref{visualize-gw}. Based on this observation, we first mine the representation for the local geometric structures from the pretrained features of the base classes and store them as the geometric words to facilitate learning of the new classes with few examples. We then learn a geometric-aware semantic representation based on the geometric words. Specifically, the geometric-aware semantic representation is a fusion of two features: 1) \textit{Class-agnostic geometric feature} obtained by the assignment of low-level features to their most similar geometric words. 2) \textit{Class-specific semantic feature} which is the output of a feature extractor.  Intuitively, our geometric-aware semantic representation allows the encoding of the transferable geometric information across classes while preserving the semantic information for effective segmentation. 

\ytComment{
We further introduce \textbf{geometric prototype} (GP) to supplement the original semantic prototype in the prototype learning.
% Although GWs are class agnostic, the frequency histogram of GWs appearing in each class is able to uniquely represent its class from the global geometric perspective. Thus, we name the histogram as the geometric prototype.
Particularly, the geometric prototype refers to the frequency histogram of GWs that can uniquely represent each class from the global geometric perspective despite GWs are class agnostic.
% Each geometric prototype is able to uniquely represent a class via different frequency ratios of GWs. 
% Based on GP, we propose a geometric-guided classifier re-weighting module to rectify biased predictions originating from semantic prototypes. 
We thus leverage GP to propose a geometric-guided classifier re-weighting module for the rectification of biased predictions originating from semantic prototypes.
%calibrate the biased prediction by the semantic prototypes.
Specifically, 
we first perform minor frequency pruning on the geometric prototypes to suppress noisy responses of the geometric words for each class. Subsequently, we measure the geometric matching score between each query point and the pruned geometric prototypes, and employ these scores as geometric-guided weights. By re-weighting the semantic logits with geometric-guided weights,
our final prediction is enriched with geometric information that is transferable across classes. Such transferable information helps to facilitate the 
segmentation of new classes that only have a few samples while preserving the knowledge of base classes.
Our main contributions can be summarized as:
}

\begin{table}[t]
\centering
\resizebox{\linewidth}{!}{
\begin{tabular}{c|c|ccc}
\toprule
\multirow{2}{*}{Setting} & Training Stage & \multicolumn{3}{c}{Testing Stage}          \\ \cline{2-5} 
                         & Learn $D_\text{train}^b$   & Learn $D_\text{train}^n$  & Access $D_\text{train}^b$ & Test Classes \\ \midrule
Full-3DSeg               & \Checkmark           &\XSolidBrush          & \XSolidBrush           & $C^b$               \\ 
FS-3DSeg                         & \Checkmark             & \Checkmark           &\XSolidBrush & $C^n$                            \\
GFS-3DSeg                         & \Checkmark              & \Checkmark           & \XSolidBrush            & $C^b \cup C^n$                 \\ \bottomrule
\end{tabular}
}
% \vspace{0.5mm}
\caption{\small{\textbf{The comparison of different settings}. $D_\text{train}^b$ and $D_\text{train}^n$ denotes the training dataset of base and novel classes, respectively. $D_\text{train}^b$ has sufficient training data, while $D_\text{train}^n$ only has few shots for each new class. $C^b$ and $C^n$ denotes the label set for base classes and novel classes, respectively.}}
\vspace{-2mm}
\label{setting comparison}
\end{table}

\begin{itemize}[leftmargin=0.35cm]
\setlength{\itemsep}{0pt}
    \item We are the first to study the %promising
    important generalized few-shot point cloud segmentation task, which is more practical than its counterparts of fully-supervised and few-shot setting in the dynamic testing environment.
    \item We propose geometric words to represent diverse basic geometric structures that are shared across different classes, and a geometric-aware semantic representation 
    %to learn generalizable representation. 
    that allows for generalizable knowledge encoding.
    \item We introduce geometric prototype to supplement the semantic prototype. We design a geometric-guided classifier re-weighting module comprising minor frequency pruning to 
    dynamically guide the segmentation of novel classes with geometric similarity.
    \item We conduct extensive experiments on S3DIS and ScanNet to verify the effectiveness of our method. Specifically, our method improves over the state-of-the-art FS-3DSeg method by 6 and 8 times on the 5-shot and 1-shot settings of ScanNet, respectively. 
   
\end{itemize}

% also dynamically recognize novel categories from only a few training examples (provided only at test time) while also not forgetting the base ones or requiring to be re-trained on them. 

% new classes is atipycal, small in size. 这是在解释，为什么new classes只有少量。

%------------------------------------------------------------------------

\section{Related Work}

\paragraph{Few-shot Learning.}
Few-shot learning aims at classification of novel classes by learning with only a few labeled samples during testing stage. There are mainly three types of approaches: meta-learning \cite{finn2017model, munkhdalai2017meta, nichol2018first, lee2019meta}, metric-learning \cite{snell2017prototypical, vinyals2016matching, ye2020few, oreshkin2018tadam} and transfer learning based methods \cite{tian2020rethinking, mangla2020charting}. Meta-learning based methods aim to predict a set of parameters that can be quickly adapted to new tasks. Metric-learning methods learn a mapping from images to an embedding space where images of the same class are clustered together. Transfer learning methods aim to learn a general feature representation during base class training, and then train a classifier for each new task. 

Generalized few-shot learning is first proposed in \cite{hariharan2017low}, which aims to classify both base and novel classes after learning from the few-shot training samples of novel classes during testing stage. 
% Hariharan \etal \cite{hariharan2017low} propose to hallucinate additional examples for novel classes to alleviate the issue of insufficient learning samples of novel classes. 
Gidaris \etal \cite{gidaris2018dynamic} and Qi \etal \cite{qi2018low} introduce cosine similarity between image feature and classifier weight as the classifier, which unifies the recognition of both novel and base categories. To reduce the biased learning of base classes, Gidaris \etal \cite{gidaris2018dynamic} and Ye \etal \cite{ye2021learning} improve the classifier of the novel classes by utilizing the knowledge of base classes. Gidaris \etal \cite{gidaris2018dynamic} propose few-shot classification weight generator to perform attention over the classification weight vectors of the base classes. Ye \etal \cite{ye2021learning} synthesize calibrated few-shot classifiers with a shared neural dictionary learned in the base class training stage. 
In this work, we study generalized few-shot point cloud segmentation, a more challenging task as it targets at dense 3D point-level classification. 
Instead of fusion classifier weights of novel classes with past knowledge \cite{gidaris2018dynamic,ye2021learning}, we use geometric prototype as an additional classifier weight to help calibrate the biased prediction. Moreover, we propose geometric-aware semantic representation to facilitate the representation learning for novel classes.

% few-shot learning
% 1. few-shot learning review
% 2. generalized few-shot learning:
% (1)low-shot hallucinate: 第一个提出， hallucinate additional examples for novel classes. Instead, we mine transferable knowledge to enhance the representation of novel classes.
% (2) dynamic and low-shot: redesign the classifier as the cosine similarity between ...  
% (3) dynamic propose explicitly exploits the
% acquired past knowledge about the visual world by incorporating an attention mechanism over the classification weight
% vectors of the base categories. similar classsifier enhancement methed is proposed in acatsle.
% In this work, we study generalized xxx, a more challenging task as it targets dense point-level classification with 3D information. instead of fusion classifier weights or generating more training samples, we utilize geometric words as the exclusive transferable knowledge in 3D to enhance the representation, and geometric prototype to enhance the discriminativeness of the classifier and representation.

\begin{figure*}[t]
\centering
\includegraphics[scale=0.61]{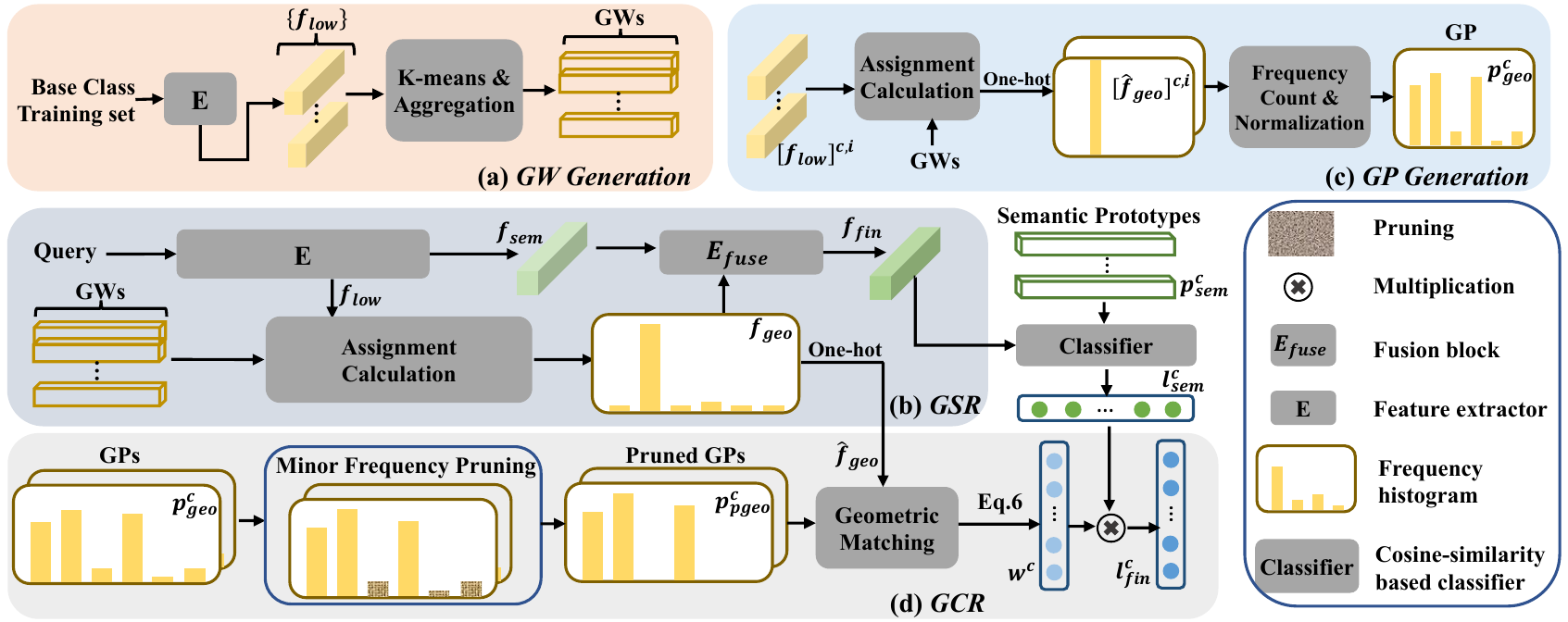}
\caption{
\ytComment{
\textbf{The overview of our proposed framework}.
\textbf{(a) GW Generation}: shows the generation of geometric words from base class training data.
\textbf{(b) Geometric-aware Semantic representation (GSR)}: semantic feature $f_\text{sem}$ is fused with geometric feature $f_\text{geo}$ as the final representation $f_\text{fin}$ for each point.
\textbf{(c) GP Generation}: shows the generation of the geometric prototype $p_\text{geo}^c$ for class c.
\textbf{(d) Geometric-guided Classifier Re-weighting (GCR)}: we compute geometric matching between geometric feature $\hat{f}_\text{geo}$ and pruned GP $p_\text{pgeo}^c$ to find potential target classes and derive weight $w^c$ to 
supplement semantic prediction $l_\text{sem}^c$. $l_\text{fin}^c$ is the final prediction logit of class c.
% The geometric-guided weight $w^c$ is multiplied with semantic prediction logits $l_\text{sem}^c$ to rectify biased predictions originating from semantic prototypes. %calibrate the biased prediction with semantic prototypes alone.
% $l_\text{fin}^c$ is the final prediction logit of class c.
}
}
\label{framework}
\vspace{-4mm}
\end{figure*}

\vspace{-4mm}
\paragraph{Few-shot Semantic Segmentation.}
Few-shot semantic segmentation performs segmentation of novel classes for images \cite{liu2020part,zhang2019canet,zhang2019pyramid,shaban2017one} or point clouds \cite{zhao2021few,lai2022tackling,mao2022bidirectional} by only learning from a few support samples. The methods for few-shot image semantic segmentation can be categorized into metric-based \cite{wang2019panet, liu2020part} and relation-based \cite{zhang2019canet, zhang2019pyramid} methods. Metric-based methods aggregate prototypes from the support set as the classifier, and perform cosine similarity with the query features. Relation-based methods concatenate the support features with query features for dense feature comparison via deep convolutional network. 
Few-shot point cloud segmentation \cite{zhao2021few} also adopts metric-based technique by performing label propagation among the query points and the multi-prototypes of each class to infer the query point label.

Generalized few-shot image semantic segmentation (GFS-2DSeg) \cite{tian2022generalized, cermelli2020prototype} aims to segment both base and novel classes in the query images during testing stage, which is more practical than the few-shot setting. CAPL \cite{tian2022generalized} leverages the contextual cues from the support set and query images to enhance the classifier weights of the base classes. %However, it ignores the under-learned novel classes and consequently shows inferior performance on novel classes.  放在实验说？
 PIFS \cite{cermelli2020prototype} finetunes on the support set of novel classes during testing stage and proposes prototype-based distillation loss to combat catastrophic forgetting of base classes. In this paper, we study generalized few-shot point cloud semantic segmentation (GFS-3DSeg), a practical yet unexplored task. Different from GFS-2DSeg which assumes the annotation of base classes is available in the novel training samples, we strictly follow the definition of GFSL that only the annotation of novel classes are provided in the support set. Consequently, both CAPL and PIFS fail to work properly under the GFS-3DSeg since both rely on the co-occurence of base classes in the support sets of novel classes to calibrate the imbalanced learning between base and novel classes. To solve this challenging problem, we propose to mine the representation for the basic geometric structures as the transferable knowledge to improve the representation and classifier of the novel classes.

\vspace{-6mm}
\paragraph{Geometric Primitives}
Geometric primitives are the fundamental components for the 3D objects, and has been initially studied in the transfer learning in 3D \cite{chowdhury2022few, zhao2022prototypical}. Chowdhury \etal \cite{chowdhury2022few} use microshapes as the basic geometric components to describe any 3D objects in the 3D object recognition task. However, object-level annotation is not available for the query point cloud in the point cloud semantic segmentation or object detection. 
Therefore, Zhao \etal \cite{zhao2022prototypical} only utilize the geometric information at point-level by enhancing the local geometric representation of each query point with a geometric memory bank. 
% Chen \etal \cite{chen2022zero} only use the assignment to the geometric primitives as the final representation for each point while ignoring the semantic information. 
Although we also performs point-level enhancement in our geometric-aware semantic representation, different from Zhao \etal \cite{zhao2022prototypical}, we inject geometric information to the high-level semantic representation to help model understand the geometric structures through learning the class semantics. Moreover, we propose to use the geometric prototype as the global geometric enhancement to each class to calibrate the biased prediction during testing stage.  

% however, object level annotation is not avaible in the point cloud of the query scene. So nipse 和zero-shot learning都只对点做了geometric enhancement. 说具体怎么做的,做的什么task.
% Although我们也对点做了enhancement， 我们是inject geometric information to the high-level represntation to help model understand the geometric compostion of the classes at the same time learning semantics. 而nipse只是enhance local geomtric repsentation of the seed points with geometric memory bank. zero-shot purely use geometric feature as the final representation of each class.
% moreover, 我们还使用了global enhancement for the classifier.

\section{Problem Formulation}
In generalized few-shot point cloud semantic segmentation, a base class training dataset $D_\text{train}^{b}$, and a novel class training dataset $D_\text{train}^{n}$ with %corresponding
non-overlapping label space $C^b \cap C^n = \varnothing$ are provided. %, respectively. 
The testing dataset $D_\text{test}$ has a label space $C_\text{test} = C^b \cup C^n$. During the training stage, the model learns the base classes $C^b$ from %sufficient
abundance of labeled point cloud data $D_\text{train}^b=\left\{\left(P_k^b, M_k^b\right)_{k=1}^{\left|D_\text{train}^b\right|}\right\}$, where $\left|D_\text{train}^b\right|$ denotes the size of $D_\text{train}^\text{b}$. Each point cloud $P_k^b \in \mathbb{R}^{m \times d_0}$ contains $m$ points with feature dimension $d_0$ and $M_k^b$ %provides
denotes the annotation of $C^b$ in $P_k^b$.

During the testing stage, the model first learns $C^n$ from limited labeled data $D_\text{train}^n = \left\{\left(P_k^{n,i}, M_k^{n,i}\right)_{k=1}^K \right\}_{i=1}^{\left|C^n\right|}$, with $K$ support point clouds per novel class $C^{n,i} \in C^n$, and $\left|C^n\right|$ is the number of novel classes.
$P_k^{n,i} \in \mathbb{R}^{m \times d_0}$ and $M_k^{n,i}$ is the binary mask indicating the presence of $C^{n,i}$. Note that during testing stage, the model does not have access to $D_\text{train}^b$.
The testing dataset $D_\text{test} = \left\{\left(P_k^q,M_k^q\right)_{k=1}^{\left|D_\text{test}\right|}\right\}$, with each query point cloud $P_k^q \in \mathbb{R}^{m\times d_0}$. $M_k^q$ represents the ground-truth annotation of $C_\text{test}$ in $P_k^q$. The goal of GFS-3DSeg is to correctly segment both $C^b$ and $C^n$ in the testing query point cloud.

% should write a brief explanation on how the semantic prototype is obtained. It should also shed some light on how the novel classes are included in the classifier.

\section{Our Method}
\paragraph{Background.}
\ytComment{
We adopt prototype learning to segment both base and novel classes during testing stage. Class prototypes are learned as the classifier weight for each class.
Base class prototypes are learned in the training stage via gradient descent, and novel class prototypes are learned by aggregating the foreground features of the support set during the testing stage. 
% The prototype captures the semantic information of each class, so we name it semantic prototype.
We name the prototype ``semantic prototype" since it captures the semantic information of each class.
The prediction of each query point is assigned by the class label of the most similar semantic prototypes.
However, naively adopting prototype learning is insufficient to learn well on new classes due to the small support set. We thus propose geometric-aware semantic representation and geometric-guided classifier re-weighting to help segmenting new classes.
% We compute cosine similarity between query point feature and semantic prototypes to find the most semantically similar class for each query point.
% how we use semantic prototypes, how it relates to ours, not to describe [25]. 
}

\vspace{-3mm}
\paragraph{Framework Overview.}
%
% \ytComment{
% As illustrated in Fig.~\ref{framework}, our proposed framework %is inspired by 
% builds upon the prototype learning method \cite{tian2022generalized} which generates semantic prototypes \footnote{We add ``semantic" in front of the prototypes used in \cite{tian2022generalized} to differentiate %with
% from our %the 
% geometric prototypes. %we propose.
% } $\left\{p_\text{sem}^c \mid c\in C_\text{test}\right\}$ as the classifier weights, and segments both base and novel classes during the testing stage. 
% The semantic prototypes of the base classes $\left\{p_\text{sem}^c \mid c\in C^b\right\}$ are obtained via gradient descent in the base class training stage, and the semantic prototypes of the novel classes $\left\{p_\text{sem}^c \mid c\in C^n\right\}$ are generated by aggregating the foreground features of the support set during the testing stage.
% We then adopt cosine similarity as the distance metric to find the most semantically similar class for each query point. 
% }
%
\ytComment{
Fig.~\ref{framework} shows the overview our framework, which consists of four main parts: a) The \textbf{geometric words (GWs)} to enhance the representation and classifier of the new classes.
%
% Based on GWs, geometric-aware semantic representation (GSR) is proposed to learn transferable representation during base class training stage.
%
b) The \textbf{geometric-aware semantic representation (GSR)} based on the GWs to learn a transferable representation during the base class training stage.
we first get the geometric feature of a point as the assignment of the GWs to the point. %describe its geometric property, and the final representation of each point is obtained by the fusion of the 
We then fuse the geometric feature with its corresponding semantic feature to get the final GSR. 
c) The \textbf{geometric prototype (GP)} to supplement the semantic prototype learned from the insufficient training samples. Specifically, 
a GP is the frequency histogram of the GWs assigned to the points in each class %and is able to
that can uniquely describe the class from the global geometric perspective.
 % We represent the geometric feature of each point as the assignment to the most similar GW to describe its geometric property, and fuse it with corresponding semantic feature as the final representation of each point
% The geometric featuren $f_\text{geo}$ is fused with corresponding semantic feature $f_\text{sem}$ as the final representation $f_\text{fin}$ of each point. 
% GSR facilitates learning the semantics of each class by recognizing the shared geometric components.
d) The \textbf{geometric-guided classifier re-weighting (GCR)} based on GPs to provide prior knowledge of each query point belonging to the potential target classes based on the geometric similarity. 
% Based on the GP, geometric-guided classifier re-weighting (GCR) is designed to provide prior knowledge of each query point belonging to the potential target classes based on the geometric similarity.
% Then, the derived geometric-guided weight $w^c$ is multiplied with semantic prediction logits $l_\text{sem}^c$ to rectify biased predictions originating from semantic prototypes. 
}
% $l_\text{fin}^c$ is the final prediction logit of class c.

% To enhance the representation and classifier of new classes, we propose  geometric words (GWs) as the transferable knowledge.
% Specifically, we propose geometric-aware semantic representation (GSR) to learn a transferable representation during base class training stage. We represent the geometric feature of each point as the assignment to the most similar GW to describe its geometric property, and fuse it with corresponding semantic feature as the final representation of each point. GSR facilitates learning the semantics of each class by recognizing the shared geometric components.
% During testing stage, we use the histogram of GWs in each class as
% the geometric prototype (GP) to supplement the original semantic prototype learned from the insufficient training samples. The GP is able to uniquely represent the corresponding class from the global geometric perspective. Based on the GP, a geometric-guided classifier re-weighting (GCR) module is designed to provide prior knowledge of each query point belonging to the potential target classes based on the geometric similarity.    

\begin{figure*}[t]
\centering
\includegraphics[scale=0.55]{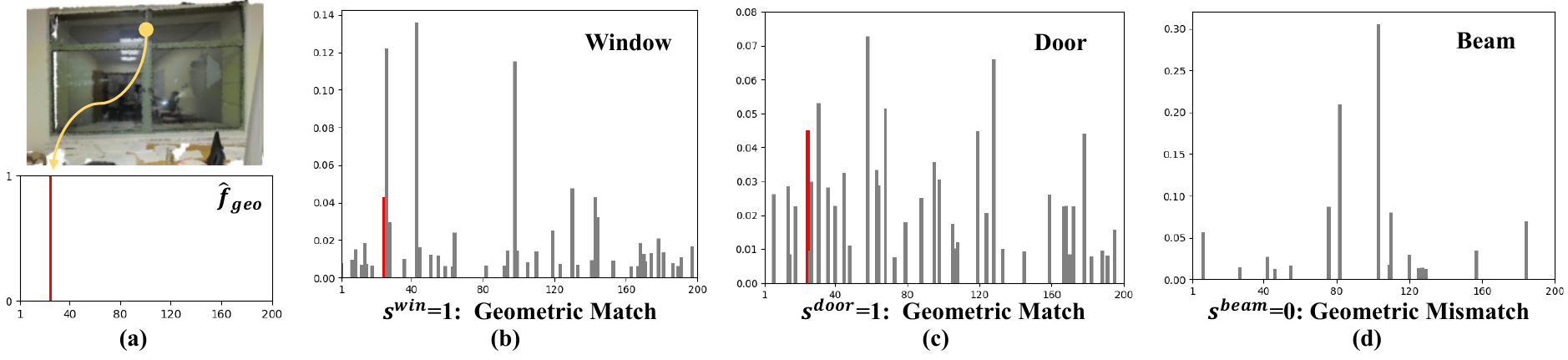}
\caption{\small{Motivation for geometric-guided classifier re-weighting. For each histogram, the horizontal axis represents the index of the GWs and the vertical axis represents the normalized frequency ratio within the class or point. (a) Visualize the geometric feature $\hat{f}_\text{geo}$ of the yellow query point on the window frame. (b),(c) and (d) shows the geometric prototypes of window, door and beam, respectively. The red bar denotes the GW with same index.}}
\label{gc}
\vspace{-4mm}
\end{figure*}

\subsection{Geometric Words}
Unlike its 2D images counterpart, 3D point clouds contain complete geometric information with shared basic geometric components. Understanding these basic geometric components helps learning across old and new classes %since they also exhibit 
due to the shared similar local geometric structures. We thus propose geometric words as the representation of these basic geometric components, and utilize it during training and testing stage to facilitate learning of new classes from few shots training point clouds. 

To obtain the geometric words, we pretrain the feature extractor $E$ of attMPTI \cite{zhao2021few} on $D_\text{train}^b$ and collect the features $\left\{f_\text{low}\right\}$ of all the points belonging to $C^b$. We concatenate the output feature of the first three EdgeConv layers and denote it as $f_\text{low} \in \mathbb{R}^{d_1}$ since lower level features contains more geometric cues. %, \ie geometric information. Then, 
We then obtain $H$ geometric words $\mathcal{G}=\left\{g_h\right\}_{h=1}^H \in \mathbb{R}^{H \times d_1}$  by applying K-means on $\left\{f_\text{low}\right\}$ to calculate the $H$ centroids. Each $g_h$ is a local aggregation of the points with similar $f_\text{low}$, \ie similar geometric characteristic.

Fig.~\ref{visualize-gw} visualizes the point-to-GWs assignments by searching points with $f_\text{low}$ that are most similar to a given geometric word $g_h$. As shown in Fig. \ref{visualize-gw}, the horizontal plane of the chair, tables and sofas are all activated to the same geometric word. It suggests that our GWs are able to represent shared geometric components among different classes. 

% 1.为什么有它
% 2.怎么计算它

% understanding the geometric stucture of each object facilitates model to capture their semantics: 1) object of the same classes usually exibit similar geometric structure. 2) different class show differen geometric strcuture. So the geometric-aware semantic representation learning helps model to learn intra-class relation and inter-class difference.

\vspace{-3mm}
\paragraph{Geometric-aware Semantic Representation.}
Based on the GWs, we propose the geometric-aware semantic representation (GSR) to enhance the feature representation of the new classes. The GSR is a fusion of a class-agnostic geometric feature $f_\text{geo} \in \mathbb{R}^H$ and a class-specific semantic feature $f_\text{sem} \in \mathbb{R}^{d_2}$. %We fuse the transferable geometric information to the semantic representation to help model understand that 
% 1) each class is the combination of basic geometric components; 2) different combinations make different classes, which in turn facilitate learning new classes. 
Specifically, the geometric feature $f_\text{geo}$ that represents the geometric information of each point is computed by the soft-assignment of its feature $f_\text{low}$ to its most similar 
GW as follow:
%
% \begin{equation}
%     f_\text{geo} = \operatorname{argmax}\left(\left[f_\text{low}\cdot g_1, \ldots, f_\text{low} \cdot g_H\right]; \text{KD=True} \right),
% \end{equation}
%
% soft assignment:
\begin{equation}
    f_\text{geo} = \operatorname{Softmax}\left(\left[f_\text{low}\cdot g_1, \ldots, f_\text{low} \cdot g_H; \tau \right] \right),
\end{equation}
%
% where $\operatorname{Softmax}$ %computes the index for the largest value of the input vector and $\text{KD}$ denotes whether to keep the input dimension in the output 
% is the softmax operator, 
where $\cdot$ denotes the cosine similarity between the feature and a geometric word. $\tau$ is the temperature to sharpen the probability vector, and we empirically set it to 10. We adopt soft assignment to make it differentiable with respect to the feature extractor. 
% Intuitively, the geometric feature of each point is a combination of the geometric words according to the similarity of its low-level feature with each geometric word. Thus, the set of geometric words forms the geometric vocabulary and each combination is the geometric document, \ie a semantic class. 
We use the final output of $E$ as the semantic feature $f_\text{sem}$.
%Then
Subsequently, $f_\text{geo}$ and $f_\text{sem}$ are concatenated and sent into a small convolution block $E_\text{fuse}$ to obtain the final representation $f_\text{final} \in \mathbb{R}^{d_3}$ for each point as follow:
 \begin{equation}
     f_\text{fin} = E_\text{fuse}\left(f_\text{geo}  \parallel  f_\text{sem}\right),
 \end{equation}
 where $ \parallel $ represents the concatenation of two vectors. 

\ytComment{
During base class training, we simulate query and fake novel class support set in each batch following \cite{tian2022generalized} to enhance the model's adaptability to unseen environments. %help model adapt to the testing environment. 
The optimization objective is to minimize cross-entropy loss computed by the prototypes generated through the assembling of $\left\{p_\text{sem}^c|c \in C^b\right\}$ and the fake novel prototypes from the simulated support set. %We refer readers to the detailed training strategy in \cite{tian2022generalized}.
We refer readers to \cite{tian2022generalized} for a more comprehensive understanding of the training strategy.
}

\subsection{\ytComment{Geometric prototype}}
Although GWs are class-agnostic, their combinations are able to represent different classes in a geometric way. 
We visualize the frequency of GWs %(\ie geometric documents) 
assigned to the points in different classes in Fig.~\ref{gc}(b), (c) and (d). The horizontal axis represents the index of the GWs and the vertical axis represents the normalized frequency ratio. 
% As shown in Fig.~\ref{gc}. the histogram for each class shows the frequency of GWs assigned to the points of the class. %by counting and normalizing the assignment to GWs of all the points within the class. 
The histogram conveys the global structure of each class via the frequency ratios of the GWs, and different classes have different histograms. 
The histogram thus uniquely represents its corresponding class and we %name 
refer to it as geometric prototype $p_\text{geo}^c \in \mathbb{R}^H$:
\vspace{-2mm}
\begin{equation}
    p_\text{geo}^c = \frac{\sum_{i=1}^{N^c} \left[\hat{f}_\text{geo}\right]^{c,i}}{N^c},
\end{equation}
where $N^c$ denotes the number of points belonging to class $c$ in the training dataset $D_\text{train}^b$ or $D_\text{train}^n$. $\hat{f}_\text{geo} \in \mathbb{R}^H$ is the hard assignment in the form of one-hot vector.
We augment the semantic prototype $p_\text{sem}^c$ with the geometric prototype $p_\text{geo}^c$, as the semantic prototype primarily encodes semantic information and, as a result, becomes insufficient in representing the new classes due to limited training samples.
%We use the geometric prototype $p_\text{geo}^c$ to supplement the original semantic prototype $p_\text{sem}^c$ that mainly aggregates the semantic information and thus is insufficient to represent the new classes alone due to lack of training samples
% and therefore would be otherwise insufficient to fully represent the new classes due to lack of training samples. 

\vspace{-3mm}
\paragraph{Geometric-guided Classifier Re-weighting.}
Based on the GP, we propose the geometric-guided classifier re-weighting module to help the prediction of the novel classes. As shown in Fig.~\ref{gc}, the corresponding geometric word for a point on the window frame is activated in the GP of window and the geometrically similar class door, but suppressed in beam which does not have the frame structure. This implies that comparing the geometric feature of the query point with GP can be employed as a hint for segmentation.
%, we can initially decide whether this point belongs to the class. 
%As shown in Fig.~\ref{gc}, a point on window frame is activated to a geometric word whose corresponding entry in $\hat{f}_\text{geo}$ is marked red. %Yet
%In contrast, this GW does not appear on the GC of beam since a beam does not have the frame structure. It indicates that this point is unlikely to belong to beam due to the geometric mismatch. 
%
% Therefore, we measure the cosine similarity between a query point's geometric feature $\hat{f}_\text{geo}$ and $p_\text{geo}^c$ to yield a geometric matching score $s^c \in \left[0,1\right]$, where $s^c=0$ means that the point does not match with class $c$ on the geometric structure. We further binarizes the $s^c$ if $s^c > 0$. The reason for binarization is to avoid bias towards GP with higher  frequency of the GW presented in $\hat{f}$
%
Therefore, we compute a geometric matching score $s^c$ based on the cosine similarity between a query point's geometric feature and GP as:
%
% Therefore, we propose the geometric matching score $s^c$ to measure the geometric similarity between query point and class $c$ as follow:
% To measure the geometric matching, we calculate the cosine similarity between GC of each class and query point feature $f_geo$ to predict the geometric matching score $s$ as follow:
\begin{equation}
    s^c = \mathds{1}\left[{p_\text{geo}^c \cdot \hat{f}_\text{geo}}\right] = \begin{cases} 1 &  p_\text{geo}^c \cdot \hat{f}_\text{geo} >0 \\ 0 & \text{otherwise} \end{cases},
\end{equation}
where $\mathds{1}[\cdot]$ is an indicator fucntion, and $s^c=1$ indicates the query point has the same geometric structure as class $c$. $c$ is the class name.

\begin{figure}[t]
\centering
\includegraphics[scale=0.67]{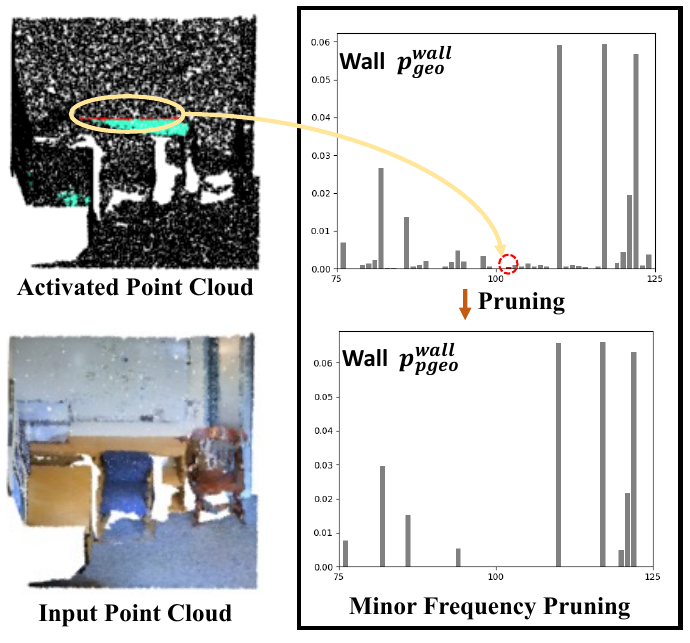}
\caption{Illustration of minor frequency pruning. The top left figure shows the green points on the table are activated to the GW representing the horizontal plane, while the red points on the vertical plane of the wall are wrongly activated to the same GW due to scene context. The right figure shows the proposed minor frequency pruning to suppress the activation to the wrong GWs introduced by scene context.}
\label{pruning}
\vspace{-2mm}
\end{figure}

% Unfortunately, the binarization also amplifies the influence of the noisy GWs from the scene context. 
% Unfortunately, the geometric matching may amplify the influence of the noisy GWs from the scene context.
% We further propose minor frequency pruning to improve the accuracy of geometric matching.

However, the geometric matching may be negatively influenced by the noisy GWs in the $p_\text{geo}^c$ due to the scene context.
As shown in Fig.~\ref{pruning}, although points on the wall (in red) are on the vertical plane, they are still activated to the same GW as the points on the horizontal table plane due to adjacency.
To suppress these noisy GWs in $p_\text{geo}^c$ and improve the accuracy of geometric matching, we propose minor frequency pruning as shown in Algorithm \ref{alg:cap}. Our motivation is that the GWs representing the typical geometric structure of the class usually contributes a large frequency on the histogram, while the GWs that are introduced by the scene context have relatively low frequencies. 
Consequently, we remove GWs corresponding to lower frequencies to only preserve the representative geometric structures. We denote the pruned geometric prototype as $p_\text{pgeo}^c$.
The frequency limit $\alpha$ in Algorithm \ref{alg:cap} denotes the amount of the frequencies to keep in the original $p_\text{geo}^c$, and the $p_\text{geo}^{c,j}$ denotes the j-th entry of $p_\text{geo}^c$.
The computation of the matching score is then updated as:
%The upgraded final matching score is then computed by:
% \begin{equation}
% \hat{s}^c:= \begin{cases} 1 & p_\text{pgeo}^c \cdot \hat{f}_\text{geo} > 0 \\ 0, & p_\text{pgeo}^c \cdot \hat{f}_\text{geo} = 0 \end{cases}.
% \end{equation}
\begin{equation}
    s^c = \mathds{1}\left[{p_\text{pgeo}^c \cdot \hat{f}_\text{geo}}\right],
\end{equation}

To highlight the geometrically matched class in the prediction, we set a weight $w^c$ to the prediction of each class according to the matching score as follow:
%
% To use the geometric matching score in the prediction, we set a weight $w^c$ to the classification logits $l_\text{sem}^c$ computed from each of the semantic prototype with the query point according to the value of $s$ as follow:
%
\begin{equation}
w^c:= \begin{cases} \beta &  s^c =1 \\ 1 & s^c=0\end{cases}.
\end{equation}
We set $\beta > 1$ to highlight the potential target classes of the query point. %We then multiply $w^c$ with the classification logit $l_\text{sem}^c$ to calculate the final prediction logit $l_\text{fin}^c$:
We then re-weight the semantic classification logit $l_\text{sem}^c$ with $w^c$ to compute the final prediction logit $l_\text{fin}^c$ as follows:
\begin{equation}
\begin{aligned}
l_\text{fin}^c = w^c \times l_\text{sem}^c,   \qquad l_\text{sem}^c = p_\text{sem}^c\cdot f_\text{fin}.
\end{aligned}
\end{equation}
The $l_\text{fin}^c$ considers both semantic and geometric similarity when segmenting new classes, which is more reliable than using semantic prediction logit $l_\text{sem}^c$ alone as shown in Tab.~\ref{s3dis gw+gc}. 
Finally, we predict the label $y$ for each query point as follow:
\begin{equation}
    y = \operatorname{argmax}\left(\operatorname{Softmax} \left(\left[l_\text{fin}^1, ..., l_\text{fin}^{\left|C_\text{test}\right|} \right]; \tau \right) \right).
\end{equation}

\begin{algorithm}[t]
\renewcommand{\algorithmicrequire}{\textbf{Input:}}
\renewcommand{\algorithmicensure}{\textbf{Output:}}
\caption{Minor Frequency Pruning}\label{alg:cap}
\begin{algorithmic}[1]
\Require Geometric prototype $p_\text{geo}^c$, frequency limit $\alpha \in \left[0,1\right]$ 
\Ensure Pruned geometric prototype $p_\text{pgeo}^c$

\State $ \gamma \gets 0$ \Comment{ $\gamma$ records the accumulated frequencies and is initialized to 0}
\State $p_\text{pgeo}^c \gets \left[0,...,0\right]^H$ \Comment{Initialize all entries of $p_\text{pgeo}^c$ to 0}

\State $\left\{\text{idx}_1, ..., \text{idx}_H \right\} \gets$ 
Sort\_Descending $\left\{p_\text{geo}^{c,j} \right\}_{j=1}^H$ \Comment{Get indices of $\left\{p_\text{geo}^{c,j} \right\}_{j=1}^H$ in descending order}

\State $\text{idx}_i \gets \text{idx}_1$

\While{$ \gamma < \alpha $}
\State $p_{\text{pgeo}}^{c,\text{idx}_i} \gets p_\text{geo}^{c,\text{idx}_i}$
\State $\gamma \gets \gamma + p_\text{geo}^{c,\text{idx}_i}$
\State $\text{idx}_i \gets \text{idx}_{i+1}$
\EndWhile
\State $p_\text{pgeo}^c \gets p_\text{pgeo}^c / \text{Sum} \left(p_\text{pgeo}^{c}\right)$ \Comment{Normalization}

% \State \Return $J_\text{rmh}$
\end{algorithmic}
\end{algorithm}

\begin{table*}[t]
\centering
\resizebox{0.98\linewidth}{!}{
% \begin{tabular}{c|cccc|cccc}
\begin{tabular}{p{2.5cm}<\centering |p{1.5cm}<\centering p{1.5cm}<\centering p{1.5cm}<\centering p{1.5cm}<\centering | p{1.5cm}<\centering p{1.5cm}<\centering p{1.5cm}<\centering p{1.5cm}<\centering}
\toprule
\multirow{2}{*}{Method} & \multicolumn{4}{c|}{5-shot} & \multicolumn{4}{c}{1-shot}   \\ \cline{2-9} 
                        & mIoU-B & mIoU-N & mIoU-A & HM  & mIoU-B & mIoU-N & mIoU-A & HM \\ \midrule
Fully Supervised                    & 76.51      & 58.69       & 68.29     & 66.42 & 76.51      & 58.69       & 68.29     & 66.42  \\ 
\midrule
attMPTI \cite{zhao2021few}          & 34.90      & 16.08       & 26.21      & 21.99    & 21.89      & 11.39       & 17.05     & 14.95\\ 
PIFS \cite{cermelli2020prototype}   & 56.99      & 19.66       & 39.76      & 29.23    & 57.85      & 14.59       & 37.88     & 23.31  \\ 
CAPL \cite{tian2022generalized}     & 73.56      & 35.18       & 55.85      & 47.51    %& 73.30      & 25.34       & 51.17     & 37.36 \\
& 72.80 & 23.87 & 50.22 & 35.67 \\
Ours                                & \textbf{73.61}      & \textbf{43.26}       & \textbf{59.60}      & \textbf{54.42} 
& \textbf{74.10} & \textbf{29.66} & \textbf{53.58} & \textbf{41.92}\\
\bottomrule
\end{tabular}
}
% \vspace{0.05mm}
\caption{Results on \textbf{S3DIS} under 5-shot and 1-shot settings.}
\label{main-s3dis}
\vspace{-2mm}
\end{table*}

\begin{table*}[t]
\centering
\resizebox{0.98\linewidth}{!}{
% \begin{tabular}{c|cccc|cccc}
\begin{tabular}{p{2.5cm}<\centering |p{1.5cm}<\centering p{1.5cm}<\centering p{1.5cm}<\centering p{1.5cm}<\centering | p{1.5cm}<\centering p{1.5cm}<\centering p{1.5cm}<\centering p{1.5cm}<\centering}
\toprule
\multirow{2}{*}{Method} & \multicolumn{4}{c|}{5-shot}   & \multicolumn{4}{c}{1-shot}   \\ \cline{2-9} 
                        & mIoU-B & mIoU-N & mIoU-A & HM & mIoU-B & mIoU-N & mIoU-A & HM \\ \midrule
Fully Supervised                    & 43.12      & 37.04       & 41.34      & 39.85    & 43.12      & 37.04       & 41.34      & 39.85   \\ 
\midrule
attMPTI \cite{zhao2021few}          & 16.31      & 3.12       & 12.35      & 5.21     & 12.97      & 1.62       & 9.57     & 2.88\\ 
PIFS \cite{cermelli2020prototype}   & 35.14      & 3.21       &25.56      & 5.88       & 35.80      & 2.54       & 25.82      & 4.75\\ 
CAPL \cite{tian2022generalized}   &38.22 & 14.39 & 31.07 & 20.88 & 38.70 & 10.59 & 30.27 & 16.53    \\
Ours           & \textbf{40.18} & \textbf{18.58} & \textbf{33.70} & \textbf{25.39}   &\textbf{40.06} &\textbf{14.78} & \textbf{32.47} & \textbf{21.55}              \\
\bottomrule
\end{tabular}
}
% \vspace{0.05mm}
\caption{Results on \textbf{ScanNet} under 5-shot and 1-shot settings.}
\label{main-scannet}
\vspace{-2mm}
\end{table*}

\section{Experiments}
\subsection{Datasets and Setup}
\paragraph{Datasets.} We evaluate on two datasets: 1) S3DIS \cite{armeni20163d} consists 272 point clouds from six areas with annotation corresponding to 13 semantic classes. We use area 6 as the testing dataset $D_\text{test}$, and leverage the other five areas to construct the training dataset for base and novel classes. 
2) ScanNet \cite{dai2017scannet} consists of 1,513 point clouds with annotation corresponding to 20 semantic classes. We use 1,201 point clouds to construct training dataset of $D_\text{train}^b$ and $D_\text{train}^n$ and the rest 312 point clouds to construct $D_\text{test}$.

For both datasets, we choose the last 6 classes with least labeled points in the corresponding dataset as the novel classes $C^n$ and the rest classes as base classes $C^b$. The motivation is to simulate the scenario in real world, where the frequency of novel class occurring is low and it is hard to collect sufficient training data. Consequently, the novel classes for S3DIS are table, window, column, beam, board and sofa. The novel classes for ScanNet are sink, toilet, bathtub, shower curtain, picture and counter. 
We follow the data pre-processing of \cite{zhao2021few} to divide each point cloud into blocks with size of $1 \ \text{meter} \times 1 \ \text{meter}$ on the $xy$ plane. From each block, we sample $m=2,048$ points as input. The dimension $d_0$ for the input feature is 9 with XYZ, RGB and normalized XYZ to the block.

\vspace{-3mm}
\paragraph{Evaluation Metrics.}
We evaluate the performance of model using mean intersection-over-union (mIoU). We use mIoU-B, mIoU-N and mIoU-A to denote the mIoU on the base classes, novel classes, and all the classes, respectively. In addition, we use harmonic mean of mIoU-B and mIoU-N to better describe the overall performance on base and novel classes, 
\ie, $\text{HM}=\frac{2 \times \text{mIoU-B}\times \text{mIoU-N}}{\text{mIoU-B}+\text{mIoU-N}}$.
% \begin{equation}
    % \text{HM}=\frac{2 \times \text{mIoU-B}\times \text{mIoU-N}}{\text{mIoU-B}+\text{mIoU-N}}.
% \end{equation}
In comparison to mIoU-A, HM is not biased towards the base classes \cite{ye2021learning}.

\subsection{Implementation details}
We adopt the feature extractor of \cite{zhao2021few} as $E$ and pretrain it on the $D_\text{train}^b$ for 100 epochs. We then perform K-means on the collection of the base class features $\left\{f_\text{low}\right\}$.  We set $H$ as 200 for S3DIS and 180 for ScanNet, respectively. 
During base class training, we perform geometric-aware semantic representation learning and learn semantic prototypes for base classes $\left\{p_\text{sem}^c \mid c \in C^b\right\}$.
% incorporate DQCE \cite{tian2022generalized} to utilize the contextual information on the query point cloud and adopt the same training strategy as \cite{tian2022generalized} to learn the base class classifier $\left\{p_\text{sem}^c|c \in C^b\right\}$.
$d_1$ and $d_2$ are both 192 following attMPTI \cite{zhao2021few}, and $d_3$ is set to 128. 
We set batch size to 32 and train for 150 epochs. 
We use Adam optimizer with initial learning rate of 0.01 and decayed by 0.5 every 50 epochs. We load the pretrained weight of the first three EdgeConv layers and set their learning rate to be 0.001. 
We compute the geometric prototypes for base classes $\left\{p_\text{pgeo}^c \mid c\in C^b\right\}$ after training completes.
During testing stage, we first obtain the semantic prototypes $\left\{p_\text{sem}^c \mid  c\in C^n\right\}$ and geometric prototypes $\left\{p_\text{pgeo}^c \mid  c\in C^n\right\}$ for novel classes by averaging $f_\text{fin}$ and $\hat{f}_\text{geo}$ (followed by minor frequency pruning) of the foreground points in the support set, respectively. We then predict the class labels for each query point via proposed GCR.
$\alpha$ is set to 0.9 and 0.95 for S3DIS and ScanNet, respectively. $\beta$ is set to 1.2.

% $d_1$ and $d_2$ are both 192 and $d_3$ is set to 128.
% We set both $\tau_1$ and $\tau_2$ to 10.

\subsection{Baselines}
We design three baselines for comparison with our method. 1) \textbf{attMPTI} \cite{zhao2021few} is the state-of-the-art FS-3DSeg method. We follow the original implementation in \cite{zhao2021few} and episodically train attMPTI on base class dataset. Upon finishing training, we collect multi-prototypes for base classes. During testing stage, we first generate multi-prototypes for novel classes from $D_\text{train}^n$. We then estimate the query label by performing label propagation among query points and prototypes of base and novel classes. 
2) \textbf{PIFS} \cite{cermelli2020prototype} is the state-of-the-art method for GFS-2DSeg that fine-tunes on $D_\text{train}^n$ to learn novel classes. We apply their proposed prototype-based distillation loss to only the scores of novel classes since we do not provide the annotation of base classes in $D_\text{train}^n$.
3) \textbf{CAPL} \cite{tian2022generalized} is the state-of-the-art method in GFS-2DSeg that performs prototype learning to learn novel classes. We remove the SCE module of CAPL since the annotations of base classes in the $D_\text{train}^n$ are not available. 
All the baselines use the same feature extractor with us for fair comparison. 

In addition, we design an oracle setting, \textbf{Fully Supervised}, where the model is trained on the fully annotated dataset of base and novel classes using the same feature extractor with us and a small segmentation head.

\begin{figure*}[t]
\centering
\includegraphics[scale=0.54]{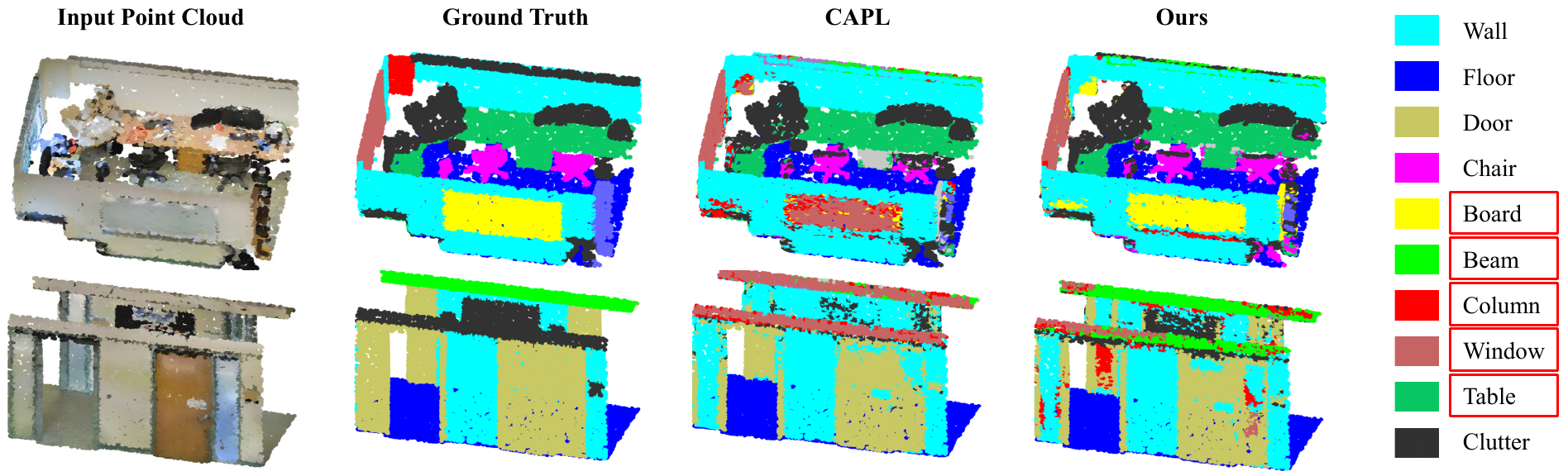}
\caption{Qualitative comparison on 5-shot setting of S3DIS dataset. Novel classes are marked with red rectangle. The target novel classes in the first row are \textbf{board}, \textbf{table}, \textbf{column} and \textbf{window}. The target novel classes in the second row is \textbf{beam}.}
\label{s3dis-seg}
\vspace{-4mm}
\end{figure*}

\subsection{Comparison with Baselines}
Tab.~\ref{main-s3dis} and Tab.~\ref{main-scannet} show the results of GFS-3DSeg on S3DIS and ScanNet, respectively. We conduct experiments in two settings with the number of support point clouds $K=\{1, 5\}$ on each dataset. We randomly generate 5 sets of $D_\text{train}^n$ using different seeds for each setting and calculate the averaged results over all 5 sets to obtain a more reliable results. 
It is clear to see that the segmentation accuracy of novel classes increases with more number of shots. Compared with all the baselines, our method is able to utilize the limited number of training samples from $D_\text{train}^n$ in a more effective way and achieves much better performance in terms of the HM and mIoU-N. 
% providing more shots for new classes achieves much better performance.
% Our method largely outperforms all the baselines in terms of the performance on novel classes and the overall performance. 

attMPTI~\cite{zhao2021few} fails to perform well on the GFS-3DSeg since it only focuses on establishing the decision boundary for those classes appearing in each episode. When including all the classes in the evaluation, the original decision boundary collapses.
PIFS~\cite{cermelli2020prototype} also performs poorly on the GFS-3DSeg. The large intra-class variances of the 3D objects make novel class adaptation difficult for the 2D fine-tuning method.
%The 2D fine-tuning method shows difficulty in learning the novel classes due to the large intra-class variance of the 3D objects.  
% on the support set of the novel classes is hard to learn a satisfactory classifier for the novel classes due to the diverse intra-class object structure. 
Moreover, their fine-tuning method leads to severe catastrophic forgetting of the base classes due to the absence of base class training data. 
% relies heavily on the annotations of base classes in $D_\text{train}^n$, which is not available in our setting. Thus, it also leads to a severe catastrophic forgetting on the base classes.
Our method is based on CAPL~\cite{tian2022generalized}. Compared to CAPL which mainly utilizes the context information to enhance the semantic prototypes of the base classes, we focus on enhancing the learning of the new classes by leveraging transferable knowledge GWs. As a result, our method shows great superiority over CAPL in the performance of the novel classes. This can also be verified in Fig.~\ref{s3dis-seg}, where our model is able to segment target new classes (board and table in the first row, and beam in the second row) more precisely.

% --------------------------- begin s3dis gw, gc -----------------------------------
\begin{table}[t]
\centering
\resizebox{\linewidth}{!}{
\begin{tabular}{cc|cccc}
\toprule
GSR & GCR                             & mIoU-B & mIoU-N & mIoU-A & HM \\ \midrule
 \XSolidBrush & \XSolidBrush     & 73.56  & 35.18  & 55.85  & 47.51   \\
\Checkmark & \XSolidBrush     & 73.48  & 40.10  & 58.07  & 51.81   \\ 
\Checkmark &\Checkmark  & \textbf{73.61}  & \textbf{43.26}  & \textbf{59.60}  & \textbf{54.42}   \\ 
\bottomrule
\end{tabular}
}
\caption{Effectiveness of geometric-aware semantic representation (GSR) and geometric-guided classifier re-weighting (GCR) on S3DIS.}
\label{s3dis gw+gc}
\vspace{-4mm}
\end{table}
% ------------------------- end s3dis gw, gc ---------------------------------------
\begin{table}[t]
\centering
\resizebox{\linewidth}{!}{
\begin{tabular}{c|cccc}
\toprule
Number of GWs                              & mIoU-B & mIoU-N & mIoU-A & HM \\ \midrule
$H=100$    & 72.45 &  \textbf{43.61} & 59.14 & 54.38       \\
 $H=150$    & \textbf{73.88} &  42.27 & 59.29 & 53.71      \\ 
$H=200$   &73.61 & 43.26 & \textbf{59.60} & \textbf{54.42}   \\
$H=250$   & 72.96  & 41.29 & 58.32   & 52.63\\
$H=400$   & 73.19 & 40.84 & 58.26 & 52.36 \\
\bottomrule
\end{tabular}
}
\caption{Ablation study of number of geometric words in S3DIS.}
\label{number of GWs}
\vspace{-2mm}
\end{table}

\begin{table}[t]
\centering
\resizebox{\linewidth}{!}{
\begin{tabular}{c|cccc}
\toprule
Frequency Limit   & mIoU-B & mIoU-N & mIoU-A & HM \\ \midrule
$\alpha=1$     & 73.50  & 41.05  & 58.50  & 52.59     \\
$\alpha=0.95$    &73.48 & 43.19 & 59.50 & 54.34     \\
$\alpha=0.9$ &   \textbf{73.61}  &\textbf{43.26}  & \textbf{59.60}  & \textbf{54.42}\\
$\alpha=0.85$ & 73.59 & 43.01 & 59.48 & 54.22 \\
% $\alpha=0.8$  &72.40 & 34.66 & 54.98 & 46.85   \\
\bottomrule
\end{tabular}
}
\caption{Ablation Study of frequency limit $\alpha$. $\alpha=1$ denotes model without minor frequency pruning in the GCR.}
\label{enegy-limit}
\vspace{-2mm}
\end{table}

% ------- ablation beta ---------
\begin{table}[t]
\centering
\resizebox{\linewidth}{!}{
\begin{tabular}{c|cccc}
\toprule
logits weight   & mIoU-B & mIoU-N & mIoU-A & HM \\ \midrule
$\beta=1$    & 73.48   & 40.10  & 58.07  & 51.81     \\
$\beta=1.2$  & \textbf{73.61} & \textbf{43.26} & \textbf{59.60} & \textbf{54.42} \\
$\beta=1.5$ & 73.20 & 42.25 & 58.90 & 53.51   \\
\bottomrule
\end{tabular}
}
\caption{Ablation Study of logits weight $\beta$. $\beta=1$ is equivalent to the model without using GCR.}
\label{logit-weight}
\vspace{-2mm}
\end{table}
% --------- end ablation beta -----

% --------- begin sota feature extractor -----
\begin{table}[t]
\centering
\resizebox{\linewidth}{!}{
\begin{tabular}{c|cccc}
\hline
Method       & mIoU-B & mIoU-N & mIoU-A & HM \\ \hline
 PTv2+CAPL    &86.35  &27.58   &59.22  & 41.68       \\ 
 PTv2+Ours   &\textbf{86.38} &\textbf{37.26}  &\textbf{63.71}  &\textbf{52.01}    \\
\hline
\end{tabular}
}
\caption{Comparison using state-of-the-art feature extractor Point Transformer v2 (PTv2) \cite{wu2022point}.}
\label{sota backbone}
\vspace{-4mm}
\end{table}
% -------- end sota feature etractor -----

\begin{table*}[t]
\begin{minipage}{0.7\linewidth}
% \vspace{1mm}
\centering
\setlength{\tabcolsep}{1pt}
\resizebox{\linewidth}{!}{
\begin{tabular}{c|ccccccccccccc}
\hline
Method & ceiling        & floor          & wall           & \textcolor{red}{beam}           & \textcolor{red}{column}         & \textcolor{red}{window}         & door           & \textcolor{red}{table}          & chair          & \textcolor{red}{sofa}           & bookcase       & \textcolor{red}{board}          & clutter        \\ \hline
CAPL   & \textbf{93.17} & \textbf{97.36} & \textbf{72.92} & 52.44          & 28.82          & 35.12          & \textbf{77.57} & 60.60          & 61.29          & 10.38          & 54.42          & 23.75          & 58.19          \\
Ours   & 92.34          & 97.16          & 70.73          & \textbf{60.37} & \textbf{32.16} & \textbf{46.30} & 76.21          & \textbf{64.41} & \textbf{63.42} & \textbf{16.99} & \textbf{54.77} & \textbf{39.30} & \textbf{60.66} \\ \hline
\end{tabular}
}
% \vspace{-3mm}
\caption{Per-class IoU. Red denotes new classes.}
\label{per-class iou}
\end{minipage}
% \hspace{0.1cm}
\begin{minipage}{0.3\linewidth}
% \vspace{-3mm}
\centering
\setlength{\tabcolsep}{1pt}
\resizebox{\linewidth}{!}{
\begin{tabular}{c|cccc}
\hline
Method                              & mIoU-B & mIoU-N & mIoU-A & HM \\ \hline
 CAPL    & 0.17 &  3.25 & 1.46 & 2.94       \\ 
Ours   &0.35 & 3.34 & 1.63 & 2.75   \\
\hline
\end{tabular}
}
\caption{Standard deviation.}
\label{std}
\end{minipage}
\vspace{-4mm}
\end{table*}

\subsection{Ablation Study}
% \vspace{-2mm}
\paragraph{Effectiveness of GSR and GCR.}
We verify the effectiveness of the geometric-aware semantic representation (GSR) and geometric-guided classifier re-weighting (GCR) on 5-shot setting using S3DIS (Tab.~\ref{s3dis gw+gc}).% and ScanNet (Tab.~\ref{scannet gw+gc}). 
The model without GSR uses $f_\text{sem}$ as the feature representation for each point. The model without GCR adopts $l_\text{sem}^c$ as the final prediction logit of class c.
Both GSR and GCR are beneficial to new class segmentation, which suggests the successful enhancement of the representation and classifier of the novel classes by using transferable geometric information.

\vspace{-3mm}
\paragraph{Analysis of the number of GWs.} 
Tab.~\ref{number of GWs} studies the influence of GW numbers on the performance of 5-shot setting using S3DIS. Although the performance varies with the different numbers of geometric words, they all largely outperform the baselines without GWs. We choose $H=200$ in our model regarding its best overall performance.

\vspace{-3mm}
\paragraph{Effectiveness of the minor frequency pruning.}
In Tab.~\ref{enegy-limit}, we analyze the effect of the minor frequency pruning in the geometric prototype. GP without pruning ($\alpha=1$) shows worst performance compared to the pruned GPs. $\alpha=0.9$ gives the best performance, thus we adopt $\alpha=0.9$ in our final model.

\vspace{-3mm}
\paragraph{Analysis of the weight $\beta$.}
We present the ablation study of the weight $\beta$ in Tab.~\ref{logit-weight}. 
$\beta = 1$ does not highlight any potential target classes predicted by geometric matching, which is equivalent to the model without GCR.
Setting a moderate threshold $>1$ is helpful to improve the segmentation performance, so we choose $\beta=1.2$ for model testing.

\vspace{-3mm}
\paragraph{Using SOTA feature extractor.}
\ytComment{Tab.~\ref{sota backbone} shows the result of replacing DGCNN-based feature extractor \cite{zhao2021few} with Point Transformer v2 (PTv2) under the 5-shot setting of S3DIS. Our method outperforms the strongest baseline CAPL by a large margin on novel classes and overall performance. It verifies that our method can work successfully %cooperate 
with a state-of-the-art point cloud feature extractor. We notice that the mIoU-N in Tab.~\ref{sota backbone} is lower than that of Tab.~\ref{main-s3dis}. 
One possible reason is that the feature extractor in \cite{zhao2021few} is specially designed to be able to quickly learn new classes from a small support set.
}

\vspace{-3mm}
\paragraph{Standard deviation and per-class IoU.}
\ytComment{Tab.~\ref{std} shows the standard deviation results on the 5 testing sets of the 5-shot setting of S3DIS. Our model shows similar variation with CAPL.
Tab.~\ref{per-class iou} shows per-class IoU of the 5-shot setting of S3DIS. Our method largely outperforms CAPL for all new classes while maintaining on-par performance on base classes. 
}

% \vspace{-3mm}
\section{Conclusion}
In this paper, we present the unexplored yet important generalized few-shot point cloud segmentation. We address the challenge of facilitating new class segmentation with limited training samples by utilizing transferable knowledge geometric words (GWs) mined from the base classes. We propose geometric-aware semantic representation to learn generalizable representation where geometric features described through GWs are fused with semantic representation. We further propose the geometric prototype (GP) to supplement the semantic prototype in the testing stage. 
Extensive experiments on two benchmark datasets demonstrate the superiority of our method. 

\vspace{-3mm}
\paragraph{Acknowledgement.} 
This research work is fully done at the National University of Singapore and 
is supported by the Agency for Science, Technology and Research (A*STAR) under its MTC Programmatic Funds (Grant No. M23L7b0021).

\vspace{5mm}
\appendix
\renewcommand\thefigure{\Alph{section}\arabic{figure}}  
\renewcommand\thetable{\Alph{section}\arabic{table}}

{\centering\section*{Supplementary Material}}

\section{Experimental Results on 3-shot Setting}
\setcounter{figure}{0}
\setcounter{table}{0}
To further validate the effectiveness of our model, we compare our method with the baselines under the 3-shot setting on S3DIS and ScanNet in Tab.~\ref{s3dis 3-shot} and Tab.~\ref{scannet 3-shot}, respectively. The results consistently illustrate that our model outperforms all baselines by a large margin on novel class segmentation, and achieves the best overall performance.

% ----------------- begin s3dis 3-shot ------------------
\begin{table}[h]
\centering
\resizebox{\linewidth}{!}{
\begin{tabular}{c|cccc}
\toprule
Methods                              & mIoU-B & mIoU-N & mIoU-A & HM \\ \midrule
 Fully Supervised    & 76.51 &  58.69 & 68.29 & 66.42      \\ 
 \midrule
 attMPTI \cite{zhao2021few} & 36.28 & 13.32 & 25.68 & 19.28 \\
 PIFS \cite{cermelli2020prototype} & 54.34 & 20.00 & 38.53 & 29.23 \\
 CAPL \cite{tian2022generalized}& \textbf{73.66} & 33.39 & 55.05 & 45.76 \\
 Ours & 73.55 & \textbf{41.55} & \textbf{58.78} & \textbf{53.04} \\
\bottomrule
\end{tabular}
}
\caption{Results on \textbf{S3DIS} under 3-shot setting.}
\label{s3dis 3-shot}
% \vspace{-4mm}
\end{table}
% ------------------ end sedis 3-shot -------------------

% ------------------ begin scannet 3-shot ---------------
\begin{table}[h]
\centering
\resizebox{\linewidth}{!}{
\begin{tabular}{c|cccc}
\toprule
Methods                              & mIoU-B & mIoU-N & mIoU-A & HM \\ \midrule
 Fully Supervised    & 43.12 &  37.04 & 41.34 & 39.85      \\ 
 \midrule
 attMPTI \cite{zhao2021few}& 16.78 & 2.42 & 12.47 & 4.24\\
 PIFS \cite{cermelli2020prototype}& 35.97 & 2.86 & 26.04 & 5.31 \\
 CAPL \cite{tian2022generalized} & 38.32 & 13.65 & 30.92 & 20.05\\
 Ours & \textbf{40.22} & \textbf{17.90} & \textbf{33.52} & \textbf{24.72}  \\
\bottomrule
\end{tabular}
}
\caption{Results on \textbf{ScanNet} under 3-shot setting.}
\label{scannet 3-shot}
% \vspace{-4mm}
\end{table}
% ------------------- end scannet 3-shot ------------------

\section{t-SNE Visualization}
\begin{figure}[h]
\centering
\includegraphics[scale=0.42]{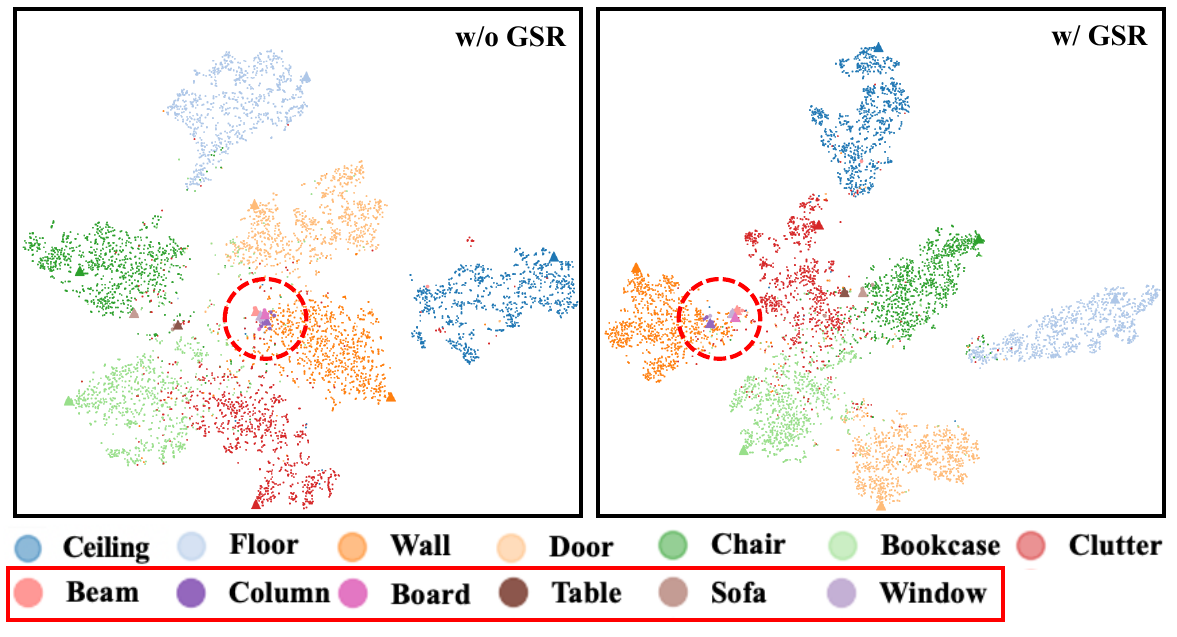}
\caption{\textbf{t-SNE visualization on 5-shot setting of S3DIS.} Small dots represent point features and triangle represents weights for semantic prototypes $P_\text{ori}$. Novel classes are indicated with the red rectangle. Best viewed zoomed-in.}
\label{tsne}
\vspace{-1mm}
\end{figure}

Fig.~\ref{tsne} displays the t-SNE visualization for S3DIS under the 5-shot setting. The difference between the left and right figures is whether the geometric-aware semantic representation (GSR) is employed. By using GSR, the representation of novel classes are more discriminative.

\section{Further Analysis on ScanNet}
To comprehensively evaluate the performance of our framework on GFS-3DSeg, we provide further analysis on ScanNet in this section.

\subsection{Ablation Study}
Tab.~\ref{scannet gw+gc} shows the ablation study on ScanNet. Both geometric-aware semantic representation (GSR) and geometric-guided re-weighting (GRW) are beneficial to novel class generalization, and our full model with both GSR and GRW performs the best regarding overall segmentation accuracy.

\subsection{Qualitative Results}
\begin{figure*}[t]
\centering
\includegraphics[scale=0.50]{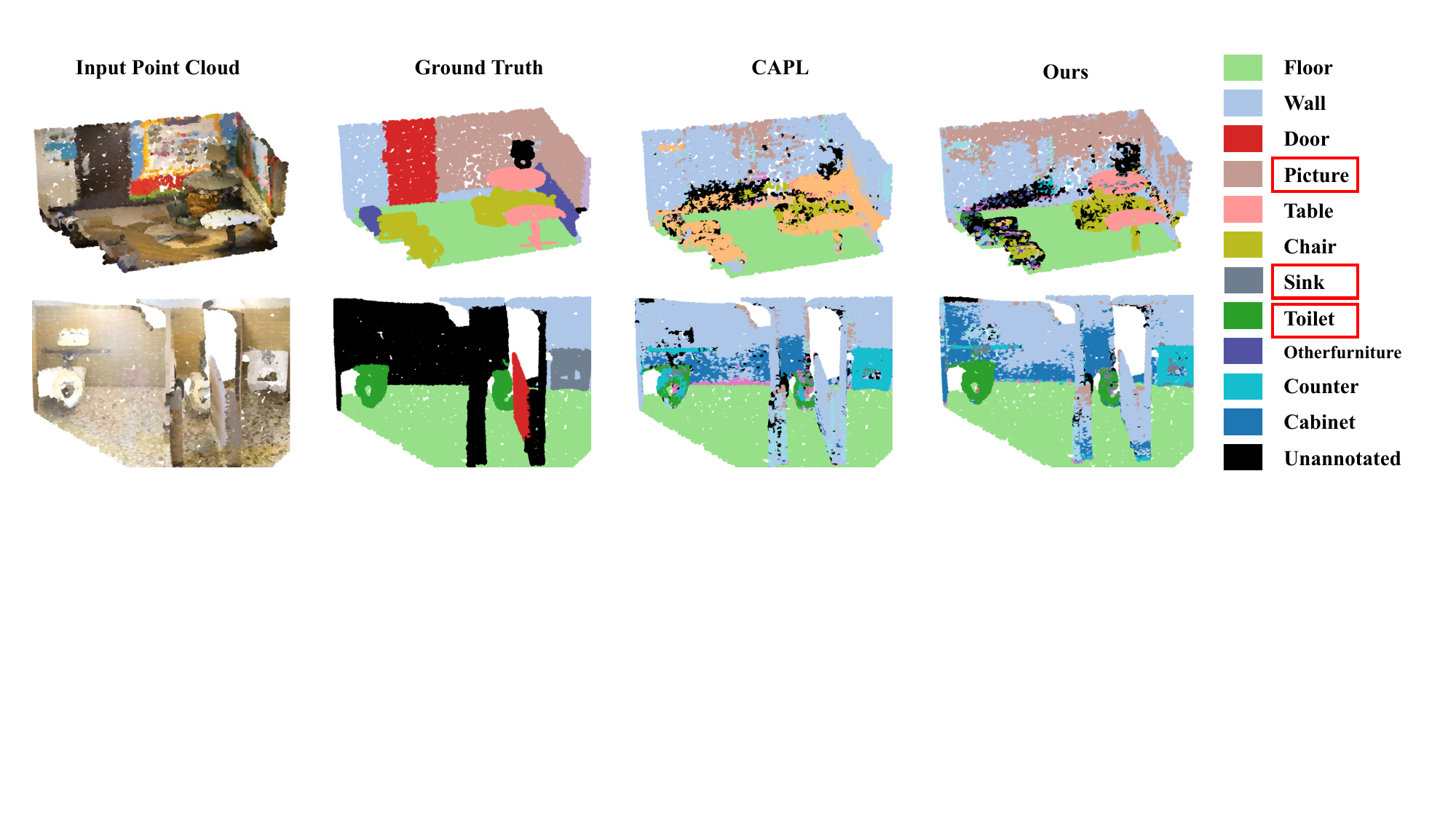}
\caption{\textbf{Qualitative comparison on 5-shot setting of ScanNet}. Target novel classes are marked with red rectangles. The target novel class in the first row is \textbf{picture}. The target novel classes in the second row are \textbf{toilet} and \textbf{sink}.}
\label{scannet-seg}
% \vspace{10mm}
\end{figure*}

The qualitative results in Fig.~\ref{scannet-seg} demonstrate that our model can segment novel classes (Picture in the first row, Toilet and Sink in the second row) more precisely than CAPL \cite{tian2022generalized}. %In the meantime
% Meanwhile, we can also perform better in base class segmentation (Table in the first row).
%Meanwhile
Concurrently, we can still maintain good segmentation performance on base classes.

% --------------------------- begin scannet gw, gc ---------------------------------
\begin{table}[t]
\centering
\resizebox{\linewidth}{!}{
\begin{tabular}{cc|cccc}
\hline
GSR & GRW                               & mIoU-B & mIoU-N & mIoU-A & HM \\ \midrule
\XSolidBrush & \XSolidBrush     & 38.22  & 14.39  & 31.07  & 20.88   \\
\Checkmark & \XSolidBrush     & \textbf{40.21}  & 17.54  & 33.40  & 24.39   \\ 
\Checkmark &\Checkmark & 40.18  & \textbf{18.58}  & \textbf{33.70}  & \textbf{25.39}   \\ 
\bottomrule
\end{tabular}
}
\caption{Effectiveness of geometric-aware semantic representation (GSR) and geometric-guided re-weighting (GRW) on ScanNet.}
\label{scannet gw+gc}
% \vspace{-4mm}
\end{table}
% -------------------------- end scannet gw, gc ----------------------------------------

% \vspace{-2mm}
\subsection{Visualization of Geometric Words}
\begin{figure*}[t]
\centering
\includegraphics[scale=0.54]{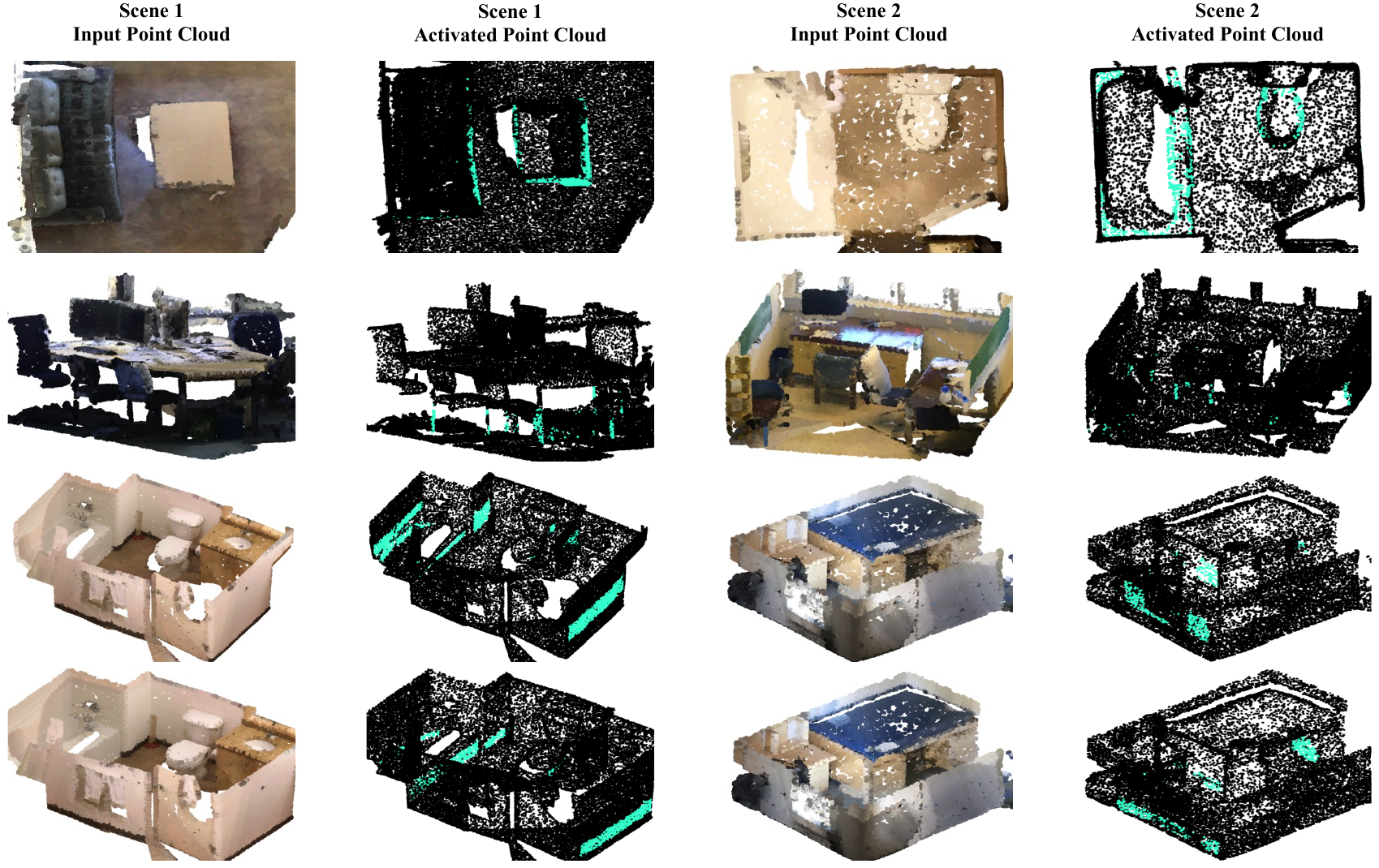}
\caption{\textbf{Visualization of geometric words on ScanNet}. Each row shows the activated point cloud regarding the same geometric word in two different scenes. The activated points are colored green.}
\label{scannet-gw}
% \vspace{-3mm}
\end{figure*}

Fig.~\ref{scannet-gw} visualizes the geometric words (GWs) on ScanNet. Each row shows two activated point clouds regarding to the same geometric word in different scenes. In the first row, the edge of the sofa, table, bathtub and toilet are all activated when provided with the same GW. In the second row, the stick of chair and table are activated. It suggests that the GWs are able to represent
shared geometric components \textbf{across different scenes and different classes}. Interestingly, we also find that GWs are height-aware. The activated parts in the third and fourth rows regarding two GWs represent vertical planes of different heights.

%Although the GWs in the third and forth rows both represent vertical planes of bathtub, wall, cabinet and bed, they are of different height.

{\small
\bibliographystyle{ieee_fullname}
\bibliography{egpaper_final}
}

\end{document}